# Structured Task Solving via Modular Embodied Intelligence: A Case Study on Rubik's Cube

Chongshan Fan[1,2], Shenghai Yuan[3,*]

*Abstract*—This paper presents Auto-RubikAI, a modular autonomous planning framework that integrates a symbolic Knowledge Base (KB), a vision-language model (VLM), and a large language model (LLM) to solve structured manipulation tasks exemplified by Rubik's Cube restoration. Unlike traditional robot systems based on predefined scripts, or modern approaches relying on pretrained networks and large-scale demonstration data, Auto-RubikAI enables interpretable, multi-step task execution with minimal data requirements and no prior demonstrations. The proposed system employs a KB module to solve group-theoretic restoration steps, overcoming LLMs' limitations in symbolic reasoning. A VLM parses RGB-D input to construct a semantic 3D scene representation, while the LLM generates structured robotic control code via prompt chaining. This tri-module architecture enables robust performance under spatial uncertainty. We deploy Auto-RubikAI in both simulation and real-world settings using a 7-DOF robotic arm, demonstrating effective Sim-to-Real adaptation without retraining. Experiments show a 79% end-to-end task success rate across randomized configurations. Compared to CFOP, DeepCubeA, and Two-Phase baselines, our KB-enhanced method reduces average solution steps while maintaining interpretability and safety. Auto-RubikAI provides a cost-efficient, modular foundation for embodied task planning in smart manufacturing, robotics education, and autonomous execution scenarios. *Code, prompts, and hardware modules will be released upon publication.*

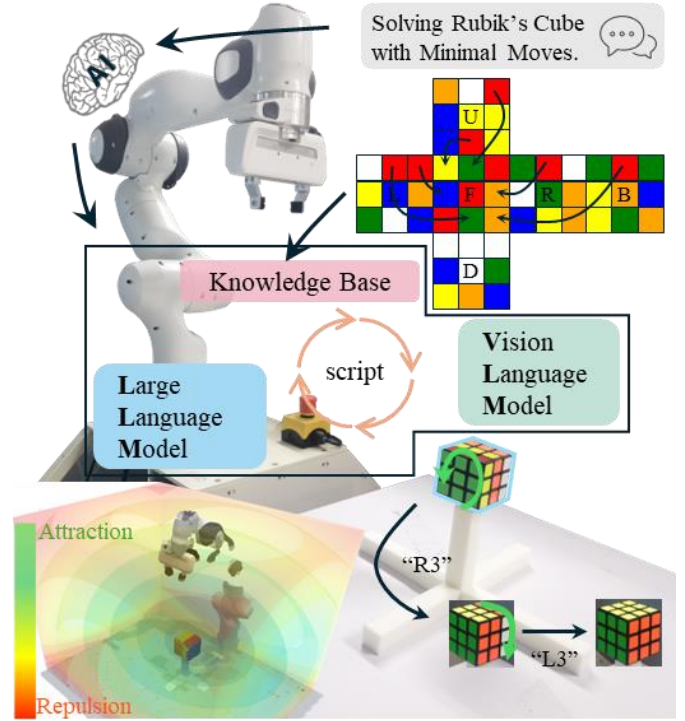

Fig. 1. The proposed Auto-RubikAI method integrates LLM and VLMs, while also introducing a targeted Knowledge Base module designed for the unique scenarios of Rubik's Cube restoration. Auto-RubikAI offers a solution approach for applying embodied intelligence methods to specialized problems.

## I. INTRODUCTION

Embodied Agents [1] has emerged as a core research direction in robotics and artificial intelligence, enabling agents to interact with physical environments using multimodal perception and action systems. However, despite recent progress in perception and action systems [2], [3], embodied agents still struggle with high-level reasoning tasks that require logical abstraction, symbolic planning, or mathematical formulation. These challenges are particularly evident in tasks such as Rubik's Cube restoration, which involve not only precise manipulation but also deep structural reasoning [4].

**Existing** embodied agents often rely on predefined rules or **handcrafted pipelines** [5], [6], limiting adaptability in unstructured environments. Recent efforts have integrated **vision-language models (VLMs)** and **large language models (LLMs)** into embodied systems [7]–[11], enabling semantic grounding and code generation. However, these systems remain limited to surface-level perception and procedural reasoning, and lack the ability to solve highly structured, **symbolic tasks** such as those involving group theory [4]. Moreover, most of these VLA-based methods depend heavily on **large-scale pretrained modules** or **massive demonstration dataset** [12], which are often impractical for tasks with tightly coupled symbolic dependencies unlimited probabilities. For example, in Rubik's Cube restoration task (see Fig. 1) where symbolic correctness and sequential execution are critical, these models often generate vague procedures or invalid actions that reveal their limitations in structured planning.

The key **challenge** lies in bridging symbolic reasoning

This work was supported by the Sustainable Development Science and Technology Special Project of Shenzhen under Grant XXXX; the Science and Technology Program of Tianjin under Grants XXXX; the National Key Research and Development Program of China under Grant XXXX; the Guang-Dong Basic and Applied Basic Research Foundation under Grant XXXX; Shenzhen Science and Technology Program (Grant XXXX); the National Natural Science Foundation of China under Grant XXXX; Beijing Tianjin Hebei Basic Research Cooperation Special Project under Grant XXXX.

[1]College of Artificial Intelligence, Nankai University, Tianjin 300071, China.
[2]Institute of Intelligence Technology and Robotic Systems, Shenzhen Research Institute of Nankai University, Shenzhen 518063, China.
[3]School of Electrical and Electronic Engineering, Nanyang Technological University, 50 Nanyang Avenue 639798, Singapore.
*Shenghai Yuan is corresponding authors. Correspondence: shyuan@ntu.edu.sg

with grounded execution, without relying on massive data or manual scripting. Rubik's Cube restoration embodies this difficulty: the task demands perceptual understanding, group-theoretic reasoning, and precise motor control, all chained in a **feedback loop**. Existing LLMs cannot autonomously generate correct symbolic steps, and VLMs lack the temporal structure for sequential planning. A solution must combine **logical abstraction**, perceptual grounding, and reliable control in a modular yet integrated framework. In the Rubik's Cube restoration task, human solvers can solve the problem by recognizing patterns and applying fixed restoration formulas along with manual operations. Existing LLMs can only provide fixed procedural steps (e.g., CFOP), which is not intu- itive enough for agents to autonomously generate restoration operation commands. Instead, more granular task planning is required. Compared to how humans achieve environmental perception and map construction through multiple real-time image captures, how can embodied intelligence leverage basic sensor inputs for environmental understanding? Furthermore, solving the Rubik's Cube restoration steps is an essential problem that must be resolved; otherwise, the task will be stuck in the first step and cannot proceed. These limitations make it difficult for current language model-based embodied intelligent systems to independently solve complex problems like Rubik's Cube restoration.

To this end, we propose **Auto-RubikAI**, a tri-modal embodied planning system that integrates a symbolic Knowledge Base (KB), a vision-language model, and a large language model. The system performs 3D scene understanding via VLM, computes optimal restoration steps through symbolic reasoning, and generates executable control code via prompt chaining. Crucially, Auto-RubikAI requires no demonstrations or retraining and generalizes well to real-world robotic exe- cution. Although restoring Rubik's Cube is not an industrial task, we argue it serves as a **proxy for symbolic manipulation tasks** in domains such as wiring, calibration, and inspection, providing a safe and interpretable benchmark for embodied symbolic planning.

**Our key contributions are as follows:**

- We propose **Auto-RubikAI**, a modular and data-free embodied planning system that integrates a symbolic Knowledge Base (KB), a Vision-Language Model (VLM), and a Large Language Model (LLM) to solve structured manipulation tasks requiring symbolic reasoning.
- We introduce a **symbolic KB module** to compute restoration steps based on group theory, bridging the reasoning gap in LLM-based planning and outperforming classical and learning-based baselines in move efficiency and interpretability.
- We develop a **retrieval-free prompt chaining pipeline** that translates symbolic plans into executable robot code without any demonstrations or retraining, enabling robust Sim-to-Real deployment on a 7-DOF arm.
- We validate Auto-RubikAI in both simulation and real-world experiments, achieving a 79% task success rate in real-world demo, and will release all code, prompts, and hardware configurations upon acceptance.

## II. RELATED WORK

The research on robotic embodied intelligence methods for solving the Rubik's Cube task in dynamic open environments mainly involves four areas: autonomous unmanned systems technology, map construction based on VLMs, task planning based on LLMs, and symbolic policy method for Rubik's Cube restoration steps.

### A. Autonomous Unmanned Systems Technology

In the past decade, autonomous robotic technologies have transitioned from classical model-based planning to data-driven and physically aware learning frameworks. Early methods primarily relied on algorithmic motion planning techniques, such as Rapidly-exploring Random Trees (RRT*) and Model Predictive Control (MPC), which performed excellently in deterministic environments but struggled in tasks characterized by dynamic uncertainty and rich contact. Recent advancements in Deep Reinforcement Learning (DRL) have addressed these limitations by enabling robots to learn complex strategies through trial-and-error interactions. For instance, Haarnoja et al. introduced the Soft Actor-Critic (SAC) algorithm [13] within the maximum entropy RL framework, significantly improving sample efficiency and exploration robustness in continuous control domains. The parallel development of imitation learning bridges the gap between human expertise and machine autonomy. Mandlekar [14] systematically analyzed how the quality of demonstrations affects policy learning, while Shafiullah [15] utilized the Transformer architecture to model long-term behavior using offline trajectory datasets. The application of robotic simulation to real-world scenarios is also crucial. Handa [16] studied Sim2Real transfer techniques that enable anthropomorphic robotic hands to master dexterous manipulation by training in random simulated environments before zero-shot deployment.

Modern robotic autonomous systems increasingly require the close integration of multimodal sensing and cognitive reasoning to operate within unstructured environments. Early unimodal approaches, such as pure visual navigation systems, have proven fragile in the face of occlusion or sensor noise. Lynch [17] pioneered language-conditioned imitation learning, allowing robots to interpret verbal instructions while align- ing actions with visual-tactile feedback. These advancements highlight the necessity of uncertainty-aware decision-making, especially when sensor modalities yield conflicting signals. At the cognitive architecture level, researchers have shifted from flat policy networks to structured task decomposition. Chen [18] reformulated multi-task reinforcement learning as a sequential modeling problem, enabling robots to dynamically prioritize subtasks in long-term missions.

In recent years, autonomous robotic technologies have developed rapidly. In mapping, Neural Radiance Fields (NeRF) [19] enhance spatial understanding by semantically reconstructing 3D environments using physical perception. In task planning, LLMs like GPT-4 [20] are being integrated into the task planning layer, allowing robots to interpret high-level goals through natural language dialogue. In trajectory generation, diffusion models are being repurposed to generate di-verse behavioral trajectories, offering probabilistic advantages over traditional deterministic planners. The leap in embodied intelligence lies in the seamless integration of perceptual foundations, physical interactions, and causal reasoning—this vision is gradually being realized through research such as Zeng's multimodal architectures [21], but there is still a long way to go toward achieving general autonomy.

### B. Application of VLMs in Robots

The application of Vision Language Models in robotics is extensive. In the areas of visual state understanding and environmental modeling, VLMs leverage their image-text alignment capabilities to transform visual inputs into semantic scene descriptions or object attribute labels. VoxPoser [22] generates a 3D value map of the scene using a large language model to guide a robotic arm in complex operations like "Open the top drawer, and watch out for that vase". RT-2 [23] utilizes the PaLI-X [24] visual language model to jointly encode images and text, directly outputting robotic action instructions such as "grab the red block". $\pi_0$ [25] is an autoregressive model based on a pre-trained VLM that incorporates action expertise and combines flow matching methods for training, allowing it to directly output action chunks. OpenVLA serves as a general visual language model, utilizing an LLM backbone and integrating a visual encoder with pre-trained features from DINOv2 [26] and SigLIP [27]. In terms of natural language interaction and task decomposition, VLMs act as an interface for "robot-environment-human", converting natural language instructions into executable actions. For instance, CLIPort [28] combines the CLIP visual-language pre-training model with a motion planner to achieve object grasping and placement driven by language instructions. Although PaLM-E [29] generates sequences of robotic actions through multimodal inputs of images and text, supporting long-horizon tasks such as "pick up the apple and put it in the fridge", it demonstrates the potential of multimodal task planning. However, its coarse-grained action generation limits its application in high-precision complex operations. Existing work [30], [31] primarily focuses on single action execution in static scenes, while tasks like Rubik's Cube require dynamic perception and multi-step closed-loop adjustments, which have yet to be thoroughly explored.

### C. Application of LLMs in Robots

In task decomposition and multi-step planning, LLMs leverage natural language understanding and symbolic reasoning capabilities to break down high-level human language instructions into executable sequences of subtasks. SayCan [32] combines the PaLM [33] language model with a robot skill library to map user instructions like "clean the table" into an action chain: "find the cup → grab → place in the sink". Code as Policies [34] utilizes large language models such as GPT-3 to generate executable Python code that directly controls robotic arms to perform tasks like stacking and assembling. Voxposer employs GPT-4 to decompose task instructions such as "put the trash in the bin" into a sequence of subtasks, including locating objects, controlling movement, grabbing, and releasing objects. Although Voxposer, SayCan, and Code as Policies demonstrate the potential of LLMs in task decomposition, they do not possess the capability to directly translate the instruction "Solve the Rubik's Cube" into specific operational steps for solving it.

In the realm of common-sense reasoning and physical law modeling, LLMs leverage pre-trained knowledge to infer strategies that adhere to physical constraints such as collision avoidance and gravitational effects. For instance, PaLM-E enhances physical law modeling capabilities by generating action sequences that account for object physical properties through multimodal image-text inputs. However, its generated instructions still lack quantitative constraints for high-precision operational parameters like rotation angles. Existing LLM-based approaches, including SayCan [32] and PaLM-E, primarily focus on semantic-level task planning, whereas Rubik's Cube restoration necessitates precise motion execution. This demands tighter integration between LLMs and motion control code. LLMs demonstrate the ability to generate code snippets compliant with programming specifications through contextual comprehension and exposure to extensive codebases. As exemplified by ProgPrompt [30], which employs programming language structures to create Python code-structured prompts based on LLM exemplars, these models can complete the code generation. Code Llama [35] achieves long code sequence generation by extending the context window.

### D. Rubik's Cube Solving

The most popular method for human Rubik's Cube restoration is CFOP [36], which has four steps for restoration. Early robot Rubik's Cube restoration mainly focused on predefined rules and precise control. For example, MIT utilized high-precision cameras to recognize colors, generated recovery steps using the Kociemba algorithm [37], and then controlled a robotic arm to execute a fixed sequence of actions. However, its robustness was limited by environmental sensitivity. Heuristic search algorithms have been employed to find optimal solutions. Korf [38] used a variant of the A* heuristic search, combined with the heuristics of the pattern database, to find the shortest possible solution.

In recent years, learning-based methods [39], [40] have attempted to address this issue through end-to-end training, but face challenges related to sample efficiency and physical constraints. For example, McAleer [39] proposed a recovery algorithm based on deep reinforcement learning, but this approach was limited to virtual environments. OpenAI [41] utilized Sim2Real technology to train a robotic hand to solve the Rubik's Cube, relying on a large amount of simulation data [42] and precise tactile sensors. The problem with learning-based methods is low sample efficiency [43]: training requires millions of simulation iterations, making it difficult to transfer to real-world applications [44]. Physical constraints are often overlooked: end-to-end policies may generate unexecutable actions (e.g., movements outside the robotic arm's workspace) [45]. Additionally, there is poor interpretability [46]: black-box models make it challenging to diagnose the sources of errors and cannot adjust policies online.

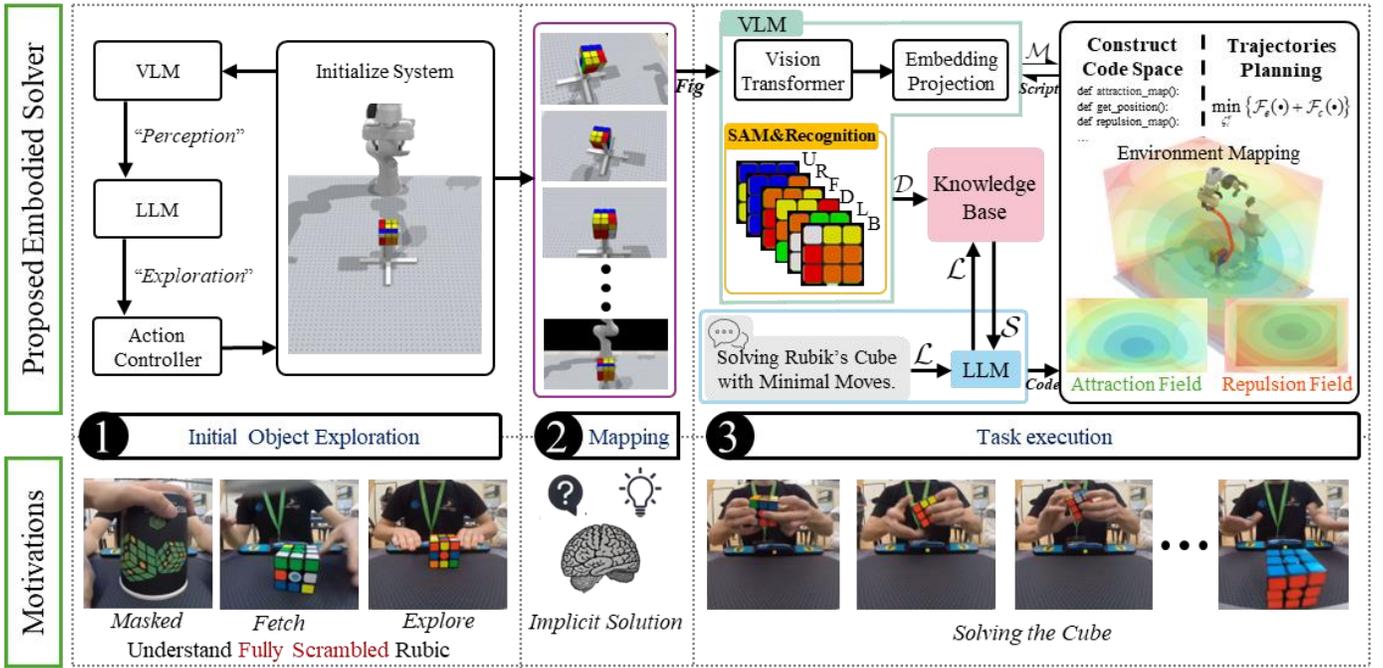

Fig. 2. Overview of the Auto-RubikAI Architecture. The VLM inputs information from the environment's RGB and depth images to create a 3D map M for environmental perception. Based on the language instruction L "Solving Rubik's Cube with Minimal Move" and the restoration steps S solved by the Knowledge Base module, Auto-RubikAI utilizes logical reasoning and code generation capabilities of the LLM to autonomously generate the corresponding control code. The code invokes trajectory planning methods and, in conjunction with the 3D map, directs robot to begin the Rubik's Cube restoration operation.

## III. PROBLEM FORMULATION

The proposed Auto-RubikAI system integrates a Knowledge Base, a vision-language model, and a large language model into a modular, closed-loop framework for autonomous task execution (see Fig. 2). Unlike traditional systems that depend on large-scale training data or rely on LLMs merely to repeat known heuristics or rule-based steps, Auto-RubikAI is entirely dataset-free and leverages the LLM to generate concrete, executable solutions tailored to the current scene and task state. The Knowledge Base provides optimal restoration logic grounded in group theory, the vision-language model constructs a structured 3D semantic map from raw observations, and the LLM synthesizes robot control code that coordinates perception and reasoning results. A trajectory generation module translates the code into motion, closing the loop between high-level understanding and low-level execution. This architecture enables embodied agents to perceive, reason, and act with precision in symbolic tasks without human demonstrations or task-specific retraining.

For convenience, we have provided a glossary in TABLE I which contains important variables of this paper.

### A. Definitions

The task of this paper is to enable the robot to autonomously generate behaviors of restoring the Rubik's Cube using an embodied intelligence approach, allowing the robot system to independently complete the Rubik's Cube task with only the human input of task instructions L.

The logical steps for executing the task are as follows: First, the robot system utilizes the VLM to understand the environ-ment and forms a character definition of the scrambled Rubik's Cube, denoted as D, which is input into the Knowledge Base module. Then, the Knowledge Base module outputs the specific steps for restoring Rubik's Cube, denoted as $\mathsf{S} = (s_{\ell_1}, s_{\ell_2}, \ldots, s_{\ell_i})$, to the LLM. The LLM notifies the predefined prompt to translate the steps into textual codes. These textual codes will invoke the VLM and optimization equations to generate the motion trajectory for each sub-step $s_{\ell_i}$. After executing the motion, the robot will repeat the execution of the sub-steps until all sub-steps are completed, ultimately achieving the overall task. After the Rubik's Cube is restored, the Knowledge Base module will be called again to check whether the cube has been restored.

TABLE I
GLOSSARY OF IMPORTANT VARIABLES

| Symbol | Space | Description |
|---|---|---|
| $\mathcal{L}$ | $\Sigma^*$ | Human Instructions by Text |
| $\mathcal{D}$ | $\Sigma^{1*54}$ | Text Definition of Scrambled Cube |
| $\mathsf{S}$ | $\Sigma^*$ | Total steps for Restoring the Cube |
| $s_{\ell_i}$ | $\Sigma^{1*2}$ | Single-step for Restoring the Cube |
| $\mathcal{M}$ | $\mathbb{R}^{w \times h \times d}$ | Environment 3D Map |
| $\mathcal{F}_e$ | $\mathbb{R}$ | Evaluation of Subtask Completion |
| $\mathcal{F}_c$ | $\mathbb{R}$ | Evaluation of Path Length |
| $\mathcal{C}$ | - | Constraints of Robot Kinematic |
| $\zeta_i$ | $\mathbb{R}^{3 \times n}$ | Discrete Waypoints of Single-step Restoration |
| $E_i$ | $E_i \in \mathcal{M}$ | Environment Map for sub-task $i$ |
| $\mathbf{q}$ | $\mathbb{R}^7$ | Robot Joint Position Vector |
| $\mathbf{v}$ | $\mathbb{R}^7$ | Robot Velocity Vector |
| $\mathbf{a}$ | $\mathbb{R}^7$ | Robot Acceleration Vector |
| $p$ | $\mathbb{R}^3$ | Discrete coordinates of Path |
| $d$ | $\mathbb{R}$ | Constrains of Euclidean Distance |

## B. Problem Definition

Consider the prompt "Solving Rubik's Cube with Minimal Move" problem presented as a free-form language instruction L. Given the complexity of this instruction L, autonomously generating the motion trajectories for the robotic arm is quite challenging. Building on the introduction of the Knowledge Base, we divide the restoration instruction S into several subtasks, combining the subtasks $(s_{\ell_1}, s_{\ell_2}, \ldots, s_{\ell_i})$ to accomplish the overall task of restoring the Rubik's Cube.

The task of autonomous trajectories generation is to create a dense sequence of waypoints $\zeta_i$ for the end effector of robot for each subtask $s_{\ell_i}$. For each subtask $s_{\ell_i}$, we define the following optimization problem:

$$\min_{\zeta_i} \{\mathcal{F}_e(E_i, s_{\ell_i}) + \mathcal{F}_c(\zeta_i)\}$$

subject to

$$\mathcal{C}(\zeta_i) = \left\{ (\mathbf{q}, \mathbf{v}, \mathbf{a}) \;\middle|\; \begin{array}{l} \forall k: q_k^{\min} \leq q_k \leq q_k^{\max} \\ \forall k: v_k^{\min} \leq v_k \leq v_k^{\max} \\ \forall k: a_k^{\min} \leq a_k \leq a_k^{\max} \end{array} \right\} \quad (1)$$

where, $\mathcal{F}_e$ represents the evaluation of the completion status for subtask $s_{\ell_i}$, and the changes in the environmental state are denoted by $E_i$. $\mathcal{F}_c$ represents the evaluation function for path length, with the expectation of completing each subtask along the shortest path. C denotes the kinematic constraints on the robot, and dynamic constraints can also be added to meet different requirements. By optimizing and solving each subtask $s_{\ell_i}$, a series of robot motion trajectories are obtained. The combination of all these trajectories collectively accomplishes all the steps S required to restore the Rubik's Cube, thereby completing the task specified by instruction L.

$$\mathcal{F}_e(E_i, s_{\ell_i}) = -\sum_{j=1}^{|\zeta_i|} \mathcal{M}(p_j) \quad (2)$$

Specifically, the expression for $\mathcal{F}_e$ is shown formula 2. Here, $\zeta_i = p_0, p_1, \ldots, p_{|\zeta_i-1|}, p_{|\zeta_i|}$ represents the trajectory of the subtask $s_{\ell_i}$, and $p_j$ denotes the coordinate representation of the trajectory, where $p_j \in \mathbb{R}^3$. Additionally, M refers to the map of the environment constructed using vision-language pairs.

$$\mathcal{F}_c(\zeta_i) = -\sum_{j=1}^{|\zeta_i|} \|p_j - p_{j-1}\|$$

subject to

$$\begin{aligned} d_{\min} &\leq \|p_j - p_{j-1}\| \leq d_{\max} \\ v_k^{\min} &\leq v_{t_j} = \frac{\|p_j - p_{j-1}\|}{t_j} \leq v_k^{\max} \\ a_k^{\min} &\leq a_{t_j} = \frac{\|v_{t_j} - v_{t_{j-1}}\|}{t_j} \leq a_k^{\max} \end{aligned} \quad (3)$$

Specifically, the expression for $\mathcal{F}_c$ is shown formula 3. The cost for the robot to complete the subtask $s_{\ell_i}$ is represented by the length of the trajectory. Additionally, the spacing, velocity, and acceleration of the trajectory are constrained by the robot's kinematic model. Furthermore, dynamic constraints can be added based on different requirements.

## IV. METHODOLOGY

In this section, we describe the specific implementation details: outlining the composition of the Knowledge Base, task decomposition and code generation based on LLM, and environment mapping based on VLM.

### A. Knowledge Base

Faced with the instruction L of "Solving Rubik's Cube with Minimal Move" presented in free-form language, existing LLMs cannot comprehend complex group theory and combinatorial mathematics problems. They are unable to directly generate instructions for rotating specific faces of the cube and instead provide solution formulas like CFOP for cube restoration. However, even with such formulaic steps, large language models still fail to understand how to generate concrete operations. To address this, we introduce a Knowledge Base module designed to generate direct operational logical steps S for restoring the Rubik's Cube.

When the robot directly faces the red surface of the Rubik's Cube, the notation B, U, F, L, D, and R respectively represent the back surface (orange), upper surface (yellow), front surface (red), left surface (blue), downward surface (white), and right surface (green). The cube contains six surfaces, each consisting of 9 colored squares. The Knowledge Base module employs a visual language model to interpret the state. Following the color sequence of central squares - yellow, green, red, white, blue, and orange - it generates a 54-character cube descriptor such as "DLLRULLFFUBBLRFLRBUDRBFUBLDRFFRDFRBLF DFBLDDUUDDRUBUURB". The encoding specification for cube representation is illustrated in Fig. 3.

Not only considering instruction L but also accounting for the robot's role as an execution carrier for restoration tasks, we aim to minimize the number of restoration steps required to solve the same scrambled Rubik's Cube. Therefore, in terms of Knowledge Base selection, we conducted comparative analyses of CFOP, DeepCubeA, and Two-Phase methodologies, with experimental results to be presented in the "Experiments" section. The Knowledge Base ultimately outputs restoration instructions formatted as "B1 U2 F2 L1 D1 R3 · · · ", where alphabetical characters denote the corresponding cube faces to rotate, and numerical values 1, 2, 3 represent clockwise rotations of 90°, 180°, and 270° respectively.

### B. Task Decomposition and Code Generation

We employ GPT-4 from the OpenAI API as the large language model to achieve task decomposition and code generation through carefully designed prompts. The LLM serves two primary functions: parsing recovery instructions S output by the Knowledge Base module and translating the parsed instructions into executable program code. A "prompt" refers to input text or instructions provided to the model to guide the LLM in generating specific types of outputs. We design multiple prompts, including global_planner prompts, code_generator prompts, interact_map prompts, and get_rotation prompts and so on. Each prompt contains multiple query examples and corresponding responses, with detailed implementations available in the open-source code repository.

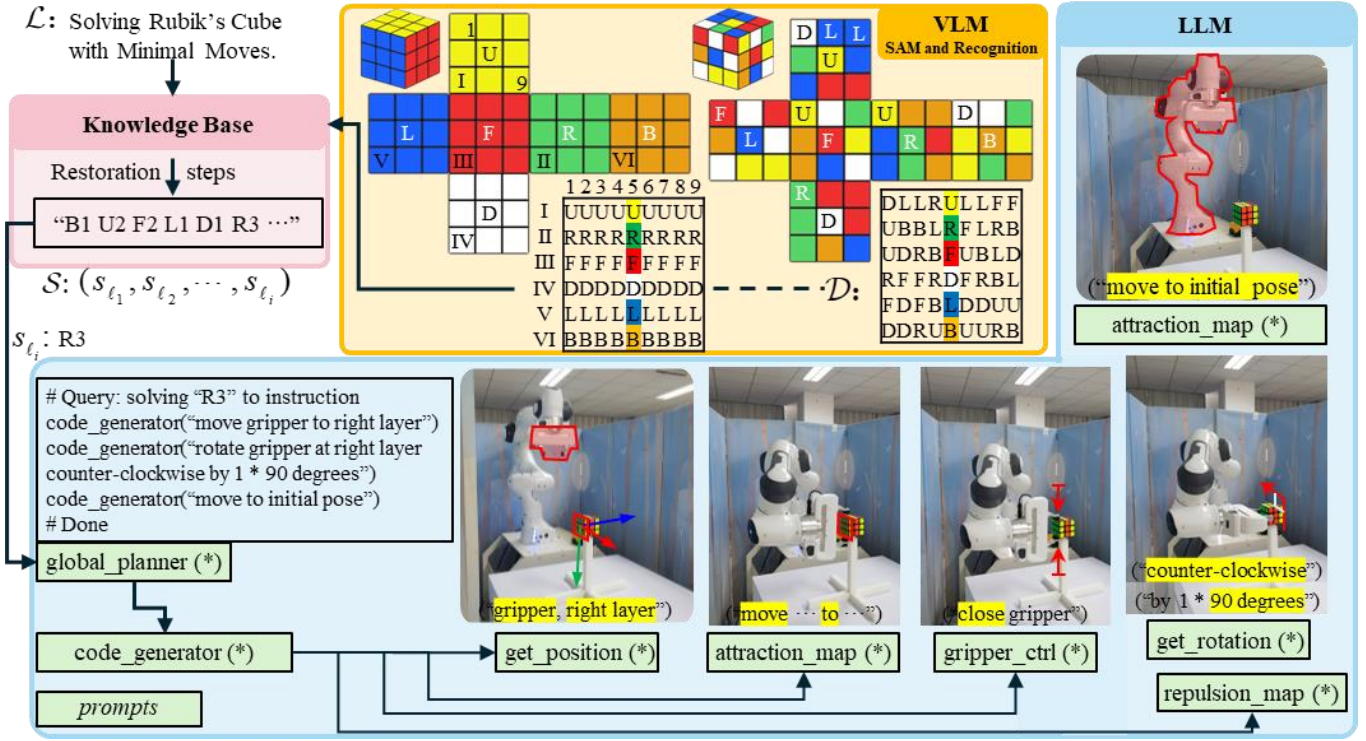

Fig. 3. Overview of the Knowledge Base Module and LLM Module. The Knowledge Base Module receives text instructions for restoring the Rubik's Cube and invokes the cube-solving method to obtain a complete restoration step instruction, such as "B1 U2 F2 L1 D1 R3 ⋯". Next, the LLM Module breaks down the complete restoration steps and converts sub-task commands like "R3" into executable code movement instructions. The conversion from "R3" to executable code movement instructions is accomplished by the LLM using multiple prompts such as "global_planner", "code_generator" and others.

When receiving a Rubik's Cube recovery instruction such as "B1 U2 F2 L1 D1 R3", the instruction $\mathcal{S}$ is decomposed into $i$ subtasks $s_\ell$, where each subtask comprises a letter-number pair. GPT-4 parses it based on examples in the global_planner prompts and generates a command script containing objects, annotations, and executable code. The complete python script generated using "R3" as an example is : "# Query: solving 'R3' to instruction; code_generator('move gripper to right layer'); code_generator('rotate gripper at right layer counter- clockwise by 1*90 degrees'); code_ generator('move to initial pose'); # Done."

Subsequently, the LLM utilizes the prompt to sequentially parse the three "code generator(*)" textual commands. This process invokes additional prompts iteratively until the generated instructions are reduced to variables or executable functions. Each command will be executed immediately after the completion of the code_ generation parsing in this round to control robotic motion. After three cycles of parsing via code_generator, the subtask "R3" is fully translated into executable instructions, and the system proceeds to the next subtask. The subtask code generation workflow for "R3" and its corresponding robotic actions are illustrated in the blue section of Fig. 3.

### C. Environment Mapping and Motion Planning

The environmental mapping process is shown in Fig. 4.. Using the 'right layer' code generated by the LLM in the previous step as the objective, and taking RGB and depth image data as input, we perform environment mapping and target localization based on a vision-language model. To balance model inference speed and accuracy, we employ OWL-ViT [47] as the open-vocabulary detector to identify the Rubik's Cube. The detected bounding boxes are then fed into Segment Anything [48] to obtain the cube's mask. Based on this mask, we localize the center colors and positions of each face of the cube. The color values are used to verify the alignment between the cube's orientation and its color configuration. The center positions, combined with map constraints, are input into the 3D point cloud to achieve environmental mapping and understanding. We define an interaction map, an ignore map, a rotation map, and a gripper control map, corresponding to the distinct language module programs interact map prompt, ignore map prompt, get rotation prompt, and gripper ctrl prompt, respectively. These collectively output the environmental 3D point cloud map. We apply Euclidean distance transform to the interact map and a Gaussian filter to the ignore map for subsequent motion planning.

In motion planning, a series of collision-free waypoints $p_j$ can be found by performing greedy shortest-distance searches on the interact map and ignore map. Rotation parameters (e.g., ensuring the gripper faces the center of each Rubik's Cube facet) are enforced at each waypoint via a rotation graph, yielding path rotation parameters in SO(3) format. Similarly, gripper open/close control parameters are enforced at each waypoint through a gripper control graph. The motion planner utilizes a total cost map, computed as a weighted sum (2:1 ratio) of the normalized interact map and ignore map, to ultimately synthesize the complete motion trajectory.

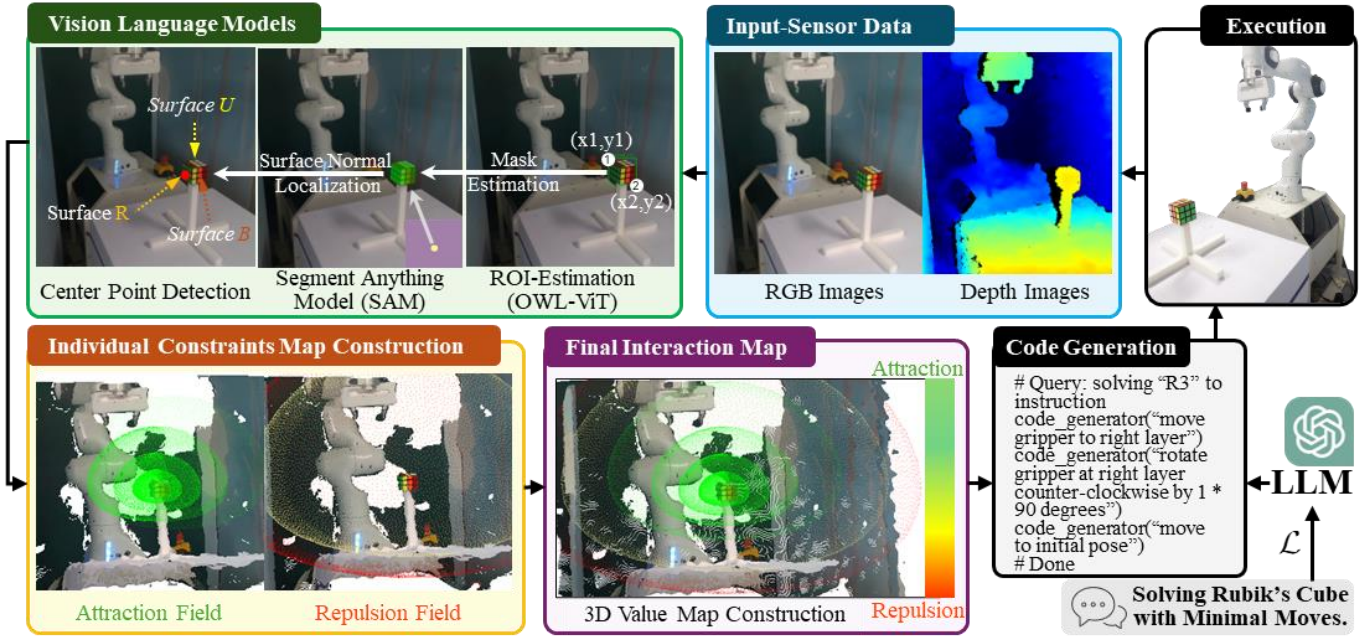

Fig. 4. Overview of closed loop reasoning for solving abstract group theory and combinatorial mathematics problems based on vision language models. Building upon sensory image data, visual language models are employed to perform detection, segmentation, recognition, and localization tasks. Furthermore, interact maps and ignore maps are systematically constructed according to predefined constraints. Ultimately, a 3D environmental region constraint map is established, thereby enhancing environmental comprehension and interaction capabilities for robotic systems.

## V. EXPERIMENT

### A. Experiment Setup

This work utilizes the Franka Emika Panda 7 degree-of-freedom robotic arm and Franka Hand gripper, along with Intel RealSense D435i RGB-D cameras to capture environmental information. The Rubik's Cube used is a standard 3x3x3 cube with a side length of 5.6 cm. A workstation equipped with an Nvidia 1080 GPU serves as the control core. Our simulation experiment is based on CoppeliaSim, the detailed code and settings can be found on our site.

### B. Baseline Selections

To evaluate the efficiency of the proposed Knowledge Base module—particularly in terms of the number of moves required to solve a given scrambled Rubik's Cube—we compare three representative baseline methods: the commonly used CFOP method [36], the two-phase method [37], and DeepCubeA [40], which is based on deep reinforcement learning. The selected baselines span both classical rule-based and learning-based approaches.

The CFOP [36], also known as the Fridrich method, solves the 3×3×3 Rubik's Cube through four sequential stages: Cross, F2L (First Two Layers), OLL (Orientation of the Last Layer), and PLL (Permutation of the Last Layer). It is widely adopted in the speedcubing community for its practical speed and consistency.

The Two-Phase [37], also known as Kociemba's algorithm, decomposes the solution process into two phases. In Phase I, the algorithm searches for move sequences that transform the cube into a constrained subspace $G_1$, in which the orientations of corner and edge pieces are fixed and the top/bottom edge pieces are relocated to their correct layers. In Phase II, the remaining permutation is solved. Multiple Phase I paths are explored to optimize for overall minimal solution length.

DeepCubeA [40] leverages approximate value iteration to train a neural network that estimates the cost-to-go from any cube state. Training is performed on scrambled states obtained by applying random inverse moves from the solved state. At inference time, the learned cost-to-go function is used as a heuristic for A* search to derive solving sequences.

Other approaches such as CubeRobot [12], which rely on closed simulation environments and datasets, are excluded from our comparison. These methods are not reproducible due to the lack of publicly available code or datasets. Similarly, methods that rely heavily on retraining with task-specific data [23], [25] are not considered here, as our system is designed to operate in a dataset-free and zero-shot manner.

### C. Evaluation Method

In terms of horizontal comparison experiments, in order to compare the recovery steps required by different algorithms, we conducted restoration experiments using randomly shuffled Rubik's Cubes. Each method was tasked with solving the same cube, allowing us to compare the steps needed for recovery. This experiment was conducted 30 times, resulting in a comparative analysis of the steps required by different methods to solve the Rubik's Cube. At the same time, we also conducted horizontal comparative experiments using human expert data. Specifically, based on the world record for the fastest human cube solving provided by www.speedcubing.com, we used the same scrambled cube and had each method solve it to compare the steps taken by humans and algorithms.

In terms of evaluating indicators for vertical comparison experiments, the following experiment was designed: The Rubik's Cube was randomly scrambled 5, 10, 20, and 40 times, and the steps required by Auto-RubikAI to solve the cube were recorded. Additionally, the accuracy of the colors on each face of the cube was assessed after completing each step.

## D. Experiment Results and Discussion

*1) Horizontal Comparison Experiment:* Table II presents the statistical results of the minimum, maximum, and average number of steps required by various methods to solve a Rubik's Cube, using 30 sets of randomly scrambled cube data. From the table, it is evident that the two-phase methods require fewer steps for restoration compared to the CFOP and DeepCubeA methods. The experiments revealed that by adjusting the parameters of the two-phase methods, it is possible to solve the cube with even fewer steps. Therefore, we adopted the modified two-phase method as the Knowledge Base for the entire task scenario. The last column of data in the table represents how much the Knowledge Base method has improved in terms of average restoration step parameters compared to the other three methods, indicating that the **Knowledge Base method outperforms the other methods**.

TABLE II
RANDOM SHUFFLE RUBIK'S CUBE
- STEPS REQUIRED BY DIFFERENT METHODS

|  | Min steps | Max steps | Avg. steps | Reduction |
|---|---|---|---|---|
| CFOP [36] | 56 | 91 | 83 | **78.3**% |
| DeepCubeA [40] | 21 | 33 | 28 | **35.7**% |
| Two-Phase [37] | 20 | 22 | 21 | **14.3**% |
| **Knowledge Base** | **17** | **19** | **18** | -% |

Table III presents the data on the fastest human cube-solving competition records[1], showing the number of steps required and average number of steps for various methods to solve the cube. In the five records, the Knowledge Base method can also complete **the cube solution below 20 steps**. The last row of data in the table still indicates that the Knowledge Base method outperforms other methods.

TABLE III
HUMAN SPEEDCUBING RECORD DATA
-STEPS REQUIRED BY DIFFERENT METHODS AND HUMANS

|  | CFOP [36] | Human | Two-Phase [37] | **Knowledge Base** |
|---|---|---|---|---|
| Record 1 | 88 | 45 | 21 | **19** |
| Record 2 | 92 | 38 | 22 | **18** |
| Record 3 | 74 | 33 | 21 | **19** |
| Record 4 | 104 | 48 | 22 | **18** |
| Record 5 | 110 | 67 | 22 | **19** |
| Avg. steps | 94 | 46 | 22 | **19** |
| Reduction | **79.8**% | **58.7**% | **13.6**% | -% |

Based on data from five of the fastest human Rubik's Cube solves, Fig. 5 shows the success rates of color matching for each of the six surfaces during the solving steps of human experts, the two-phase method, and the Knowledge Base method. The solid line represents the average color-matching rate for the six surfaces, while the dashed lines indicate the maximum and minimum average color-matching rates. From the figure, it is evident that the Knowledge Base method achieves the Rubik's Cube solution with the fewest moves. Compared to human operational methods, the Knowledge Base method exhibits greater fluctuations in color-matching rates. This experimental phenomenon suggests the computer algorithm based on symbolic policy can adopt more aggressive operational steps, allowing them to complete the Rubik's Cube solving task in fewer moves. Similarly, due to the limitations imposed by the comprehensibility of the solving algorithms and the number of formulas used, humans demonstrate a more uniform increase in color-matching rates during the experiment.

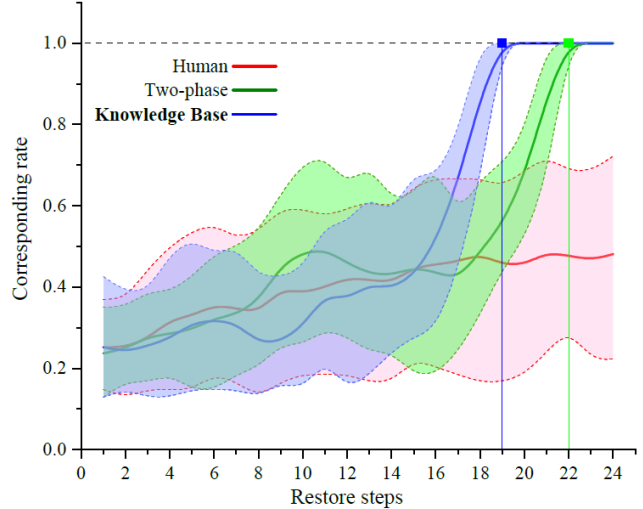

Fig. 5. The variation curves of the average color correspondence rate across six surfaces at different restoration steps, derived from five competition record datasets using various restoration methods, are illustrated in the figure. As demonstrated in the graphical representation, the Knowledge Base method exhibits superior performance compared to other approaches.

As shown in Fig. 6, "F_*color*" represents one surface of the Rubik's Cube. The horizontal axis in the figure indicates the number of rotations of the Rubik's Cube, while the vertical axis represents the color correspondence rate of that surface of the cube, used to observe changes in the color correspondence rate during the restoration process and to evaluate whether the cube has been restored. Additionally, the short horizontal lines above and below the curve represent the maximum and minimum values of the color correspondence rate during the experiment. From Fig. 6, it can be observed that when using the Knowledge Base method, the cube can be restored in 20 steps. Moreover, the trend of the curve also indicates that as the restoration of the cube approaches, the color correspondence rates of each surface become increasingly higher, aligning with the actual situation.

*2) Vertical Comparison Experiment:* We conducted the following experiments using Auto-RubikAI. Through a random scrambling program, four experimental groups were established by randomly scrambling the Rubik's Cube 10, 20, 30, and 40 times respectively. Each group was repeated 50 times for statistical analysis of the results shown in Table 4. The first column of the table indicates the experimental grouping; the second column presents the success rate of knowledge-based solving for scrambled cubes, where "(num's)" denotes the number of moves required to fully restore the cube; the third column shows the success rate of LLM-generated code for cube restoration steps $s_{\ell_i}$; and the fourth column displays the overall task success rate.

---
[1] www.speedcubing.com

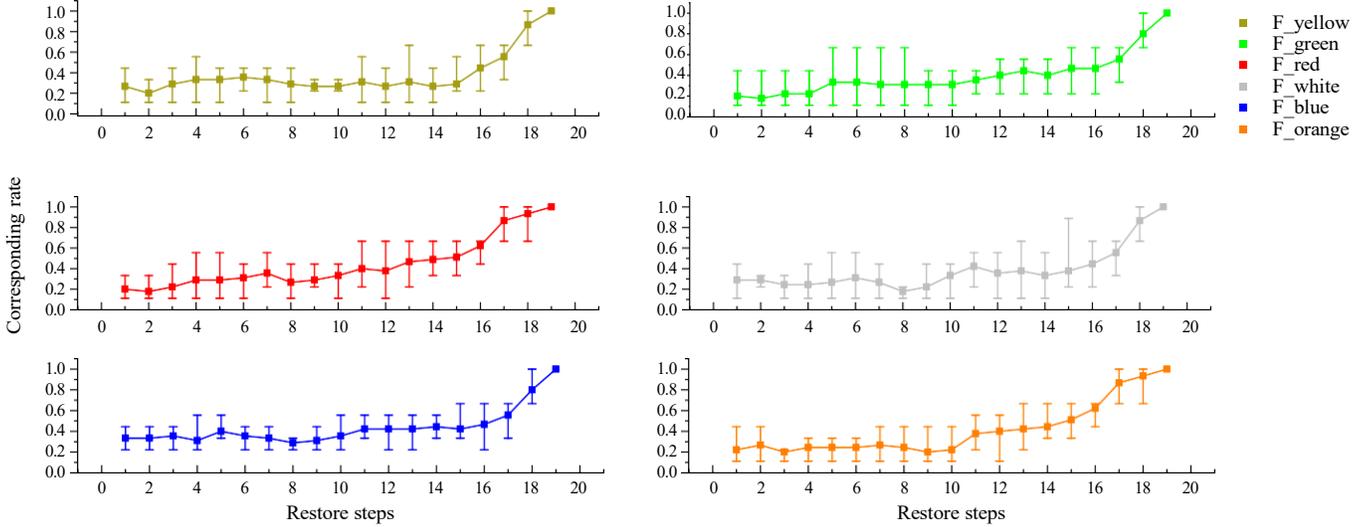

Fig. 6. The graph shows the changes in the color correspondence rates of each surface of the Rubik's Cube when using the Knowledge Base method. The trend of the color correspondence rates for each surface is similar, and **the cube can be restored in 20 steps.**

TABLE IV
STATISTICS ON THE SUCCESS RATE OF TASKS IN EACH STAGE OF AUTO-RUBIKAI

| Randomly Shuffle /times | Knowledge Base Solving Success (Restoration Steps) | LLM Code Parsing Success | Task Success |
|---|---|---|---|
| 10s | $2\%_{(6s)} 6\%_{(7s)} 16\%_{(8s)} 20\%_{(9s)} 44\%_{(10s)} \mid 12\%_{(\text{FAIL})}$ | 93.18% | 78% |
| 20s | $2\%_{(9s)} 2\%_{(13s)} 10\%_{(17s)} 40\%_{(18s)} 36\%_{(19s)} \mid 10\%_{(\text{FAIL})}$ | 88.89% | 74% |
| 30s | $8\%_{(17s)} 38\%_{(18s)} 38\%_{(19s)} 8\%_{(20s)} \mid 8\%_{(\text{FAIL})}$ | 95.65% | 84% |
| 40s | $44\%_{(18s)} 40\%_{(19s)} 6\%_{(20s)} \mid 10\%_{(\text{FAIL})}$ | 93.33% | 80% |
| Average | $90\%_{(\text{SUCC})} \mid 10\%_{(\text{FAIL})}$ | **92.76%** | **79%** |

"(*num* s)" denotes the number of steps required to restore the cube fully, e.g. (6s) requires 6 steps to restore the Rubik's Cube.

From Table IV, it can be seen that the average success rate of solving the Rubik's Cube restoration step using the knowledge base method is 90%, indicating that the cube restoration knowledge base can generate effective cube restoration steps. This also proves the feasibility of solving complex group theory and combinatorial optimization problems through special programs. The 92.76% average success rate in LLM code parsing reveals substantial potential for large language models in code generation and semantic parsing. The overall task success rate of 79% remains acceptable and serves as a valid assessment for embodied intelligence systems in **completing comprehensive task workflows encompassing environmental observation, state modeling, instruction generation, and autonomous execution**.

Figure 7 presents an **error analysis** and Knowledge Base solution analysis of the experiments. The error covers Knowledge Base (KB), LLM, and Execution (EXE) failures. Each failure rate is defined as $r_x = n_x/n_{\text{tot}}$, where $x \in \{\text{KB}, \text{LLM}, \text{EXE}\}$, $n_x$ is the number of failures in that category, and $n_{\text{tot}} = 42$ is the total number of failures. The observed rates are $r_{\text{KB}} = 47.62\%$, $r_{\text{LLM}} = 30.95\%$, and $r_{\text{EXE}} = 21.43\%$. Specifically, KB errors are further divided into color recognition errors and solution timeout errors. LLM errors encompass annotation errors and code generation errors. EXE errors include target localization errors, motion errors, and initial state errors. As shown in the bottom-right of Fig. 7, all four sets of experimental results indicate that the Knowl-edge Base method completes the Rubik's Cube restoration procedure within 20 steps, validating the effectiveness of the Knowledge Base method. It should be pointed out that the experimental results of 10 random scrambles in the first group demonstrated the KB method could restore the Rubik's Cube with the minimum number of steps (i.e., closest to the effective shuffle number), unconstrained by restoration formulas.

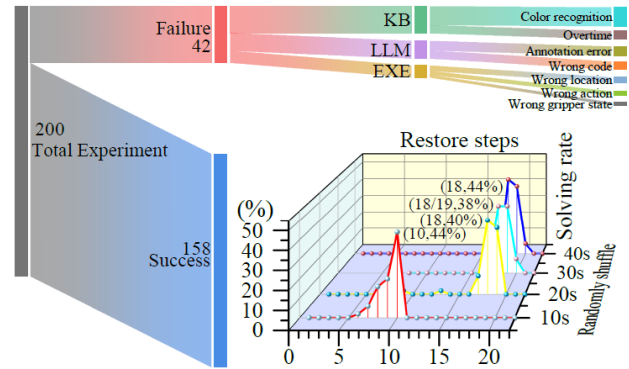

Fig. 7. Overview diagram of vertical comparative experimental analysis. Out of 200 experimental trials, 158 cases resulted in success and 42 cases resulted in failure. The causes of these errors can be categorized into three types: Knowledge Base (KB) errors, LLM errors, and execution (EXE) errors. Concurrently, the line chart in the bottom-right corner demonstrates that the selected Knowledge Base method can complete Rubik's Cube restoration procedures within 20 operational steps.

TABLE V
COMPARATIVE ANALYSIS OF EMBODIED INTELLIGENCE METHODS. FOR PRETRAINED METHODS REQUIRING DEMONSTRATIONS, HUMAN DEMO DATA WAS RECORDED AND USED FOR TRAINING.

| Method | Year | Dataset Dependency | Strategy | Success Rate |
|---|---|---|---|---|
| Voxposer [22] | CoRL2023 | No | Pre-set Prompts | Failed |
| RT-2 [23] | CoRL2023 | Yes | Behavioral Cloning | - |
| $\pi 0$ [25] | Arxiv2024 | Yes | Behavioral Cloning | - |
| **Auto-RubikAI** | - | **No** | **Knowledge Base** | **79%** |

"-" in the success rate indicates that there is no dataset to support the experiment. This further illustrates the advantages and necessity of the proposed Auto-RubikAI method.

### E. Comparative Experiments and Analysis

Considering the existing embodied AI large models such as VoxPoser [22], RT-2 [23] and $\pi 0$ [25], this section will analyze the challenges encountered when applying current embodied intelligence models to Rubik's Cube restoring tasks (see Table V), while elucidating the rationale for adopting a Knowledge Base approach in Auto-RubikAI to address these limitations. To enable embodied intelligent agents to autonomously complete the special task of Rubik's Cube restoration, this paper employs LLM for text interaction and code generation, while utilizing VLMs for environmental mapping and understanding. During the research process, we discovered that when interacting with existing GPT-4 large language models using Rubik's Cube images combined with natural language instructions, they could not provide specific restoration steps but only offered formulaic methods for cube restoration (which is insufficient for agents to complete the entire task). Therefore, introducing a dedicated module for restoration step-solving becomes essential. Additionally, we recognize that when addressing special problems characterized by sufficient complexity and scarce training data, large language models require substantial additional data for training and learning, which incurs significant costs. For such problems, the Knowledge Base module is more accurate and efficient than relying solely on data collection and model training.

When confronted with new tasks, VoxPoser utilizes LLM for semantic understanding and planning, generating new codes based on multiple predefined prompts. When Voxposer is applied to the restoration task, it fails during the LLM planning phase. This limitation stems from LLM's inherent lack of ability to solve complex group theory and combinatorial optimization problems, making it unable to provide effective recovery sequences through instruction prompts.

As evidenced in the literature, $\pi 0$ adopts a novel architecture combining pre-trained VLMs with flow matching techniques. Through pre-training on massive datasets exceeding 10,000 operational hours across diverse robots and tasks, it demonstrates remarkable generalization capabilities and physical intelligence. But, it is obvious that collecting a dataset is costly for Rubik's Cube restoration task, and this dataset is not applicable to other tasks.

The RT-2 model also requires training with multiple images, text instructions, and robotic data to achieve the generalization capability needed to adapt to diverse tasks. When applied to Rubik's Cube restoration, it faces the same limitations as VoxPoser and $\pi 0$: the inability to generate valid, step-by-step symbolic solutions based on combinatorial logic.

While RT-2 and $\pi 0$ show promise in general embodied manipulation tasks, their reliance on **massive datasets and demonstration-aligned action supervision** renders them **ill-suited for structured symbolic problems like Rubik's Cube solving**. Without retraining on **problem-specific data—which is often unavailable**—these models **fail to produce meaningful or verifiable action plans**, highlighting the **limitations of current VL-Action pretraining paradigms** in **logic** intensive domains.

In summary, existing embodied intelligence models face the challenge of failing to generate specific step-by-step solutions for Rubik's Cube restoration tasks, thereby preventing task progression. Experimental results demonstrate that the approach integrating language models with a symbolic Knowledge Base proves effective in solving Rubik's Cube problems, thus offering a promising direction for embodied intelligence in structured reasoning tasks.

## VI. CONCLUSION

In this work, we show that many existing embodied agent methods face challenges in solving the Rubik's Cube, particularly due to difficulties in generating concrete, verifiable, and executable action sequences without training data. As noted in prior studies [49], [50], **current large language models and vision-language models exhibit limitations in symbolic reasoning capacity needed for tasks requiring combinatorial logic and multi-step planning**.

To address this, we propose *Auto-RubikAI*, a modular embodied planning system that integrates a symbolic Knowledge Base (KB), a vision-language model (VLM), and a large language model (LLM). Crucially, our system is entirely **dataset-free and requires no retraining or demonstrations**, instead leveraging structured symbolic reasoning to produce executable robot code. Unlike conventional LLM-based approaches that merely recite rule to solve the cube, Auto-RubikAI grounds perception and control through explicit logic and real-world execution. We summarize three key takeaways:

- **LLMs alone are insufficient** for symbolic manipulation tasks; they tend to match or recite patterns rather than derive structured, verifiable plans.
- **Perception and language require structure to be effective**; high-level autonomy collapses without symbolic grounding.
- **Auto-RubikAI bridges the gap between perception, reasoning, and execution** by restoring structure through a symbolic Knowledge Base, enabling interpretable and robust embodied planning.

Although Rubik's Cube is not an industrial task, it serves as a rigorous proxy for structured manipulation challenges in domains such as mechanical assembly, wiring, and calibration, where symbolic correctness and sequential decision-making are essential. We believe Auto-RubikAI offers a promising direction for structured planning in embodied AI. **Code, prompts, and implementation** details will be **released** upon acceptance.